\pdfoutput=1 
\documentclass[conference]{IEEEtran}
\IEEEoverridecommandlockouts
\usepackage[singlelinecheck=false]{caption}
\def\tablename{Table}
\usepackage{amsthm}
\theoremstyle{definition}
\newtheorem{definition}{Definition}
\usepackage{cite}
\usepackage{array}
    \newcolumntype{P}[1]{>{\centering\arraybackslash}p{#1}}
    \newcolumntype{M}[1]{>{\centering\arraybackslash}m{#1}}
\usepackage{amsmath,amssymb,amsfonts}
\usepackage{algorithmic}
\usepackage{graphicx}
\usepackage{textcomp}
\usepackage{xcolor}
\def\BibTeX{{\rm B\kern-.05em{\sc i\kern-.025em b}\kern-.08em
    T\kern-.1667em\lower.7ex\hbox{E}\kern-.125emX}}
\DeclareRobustCommand{\IEEEauthorrefmark}[1]{\smash{\textsuperscript{\footnotesize #1}}}
\begin{document}

\title{Prompt Evolution for Generative AI: A Classifier-Guided Approach}

\author{
\IEEEauthorblockN{Melvin Wong\IEEEauthorrefmark{1}\textsuperscript{*}, 
Yew-Soon Ong\IEEEauthorrefmark{1,2},
Abhishek Gupta\IEEEauthorrefmark{3},
Kavitesh Kumar Bali\IEEEauthorrefmark{2},
Caishun Chen\IEEEauthorrefmark{2}}
\IEEEauthorblockA{ 
\textit{\IEEEauthorrefmark{1}School of Computer Science and Engineering (SCSE), Nanyang Technological University (NTU), Singapore} \\
\textit{\IEEEauthorrefmark{2}Centre for Frontier AI Research (CFAR), Agency for Science, Technology and Research (A*STAR), Singapore} \\
\textit{\IEEEauthorrefmark{3}Singapore Institute of Manufacturing Technology (SIMTech), Agency for Science, Technology and Research (A*STAR), Singapore} \\
\{wong1250, asysong\}@ntu.edu.sg, abhishek\textunderscore gupta@simtech.a-star.edu.sg,
\{balikk, chen\textunderscore caishun\}@cfar.a-star.edu.sg}
\thanks{*Corresponding author. \textcopyright 2023 IEEE. Personal use of this material is permitted.  Permission from IEEE must be obtained for all other uses, in any current or future media, including reprinting/republishing this material for advertising or promotional purposes, creating new collective works, for resale or redistribution to servers or lists, or reuse of any copyrighted component of this work in other works.}
}

\maketitle


\begin{abstract}
Synthesis of digital artifacts conditioned on user prompts has become an important paradigm facilitating an explosion of use cases with generative AI. However, such models often fail to connect the generated outputs and desired target concepts/preferences implied by the prompts. Current research addressing this limitation has largely focused on enhancing the prompts before output generation or improving the model's performance up front. In contrast, this paper conceptualizes prompt evolution, imparting evolutionary selection pressure and variation during the generative process to produce multiple outputs that satisfy the target concepts/preferences better. We propose a multi-objective instantiation of this broader idea that uses a multi-label image classifier-guided approach. The predicted labels from the classifiers serve as multiple objectives to optimize, with the aim of producing diversified images that meet user preferences. A novelty of our evolutionary algorithm is that the pre-trained generative model gives us implicit mutation operations, leveraging the model’s stochastic generative capability to automate the creation of Pareto-optimized images more faithful to user preferences.

\end{abstract}


\section{Introduction}
Generative AI, such as text-to-image diffusion models, has become popular for producing novel artworks conditioned on user prompts \cite{b1,b2}. A prompt is usually text expressed in free-form natural language but could also take the form of reference images containing hints of target concepts/preferences. Given such input prompts, pre-trained generative systems infer the semantic concepts in the prompts and synthesize outputs that are (hopefully) representative of the target concepts. However, a study conducted in \cite{b7} found that the generative models conditioned using prompts often failed to establish a connection between generated images and the target concepts.

Research addressing this limitation has mainly focused on enhancing the prompts before output generation \cite{b8,b10,b11,b12} or improving the model's performance beforehand \cite{b14,b17,b18}. In this paper, we conceptualize an alternative paradigm, namely, \emph{Prompt Evolution} (which we shall define in Section \ref{intropromptevo}). We craft an instantiation of this broad idea showing that more qualified images can be obtained by re-generating images from the generative model multiple times, evolving the outputs at every generation under the guidance of multi-objective selection pressure imparted by multi-label classifiers evaluating the satisfaction of user preferences. While a powerful deep generative model alone could come up with near Pareto optimal outputs, our experiments show that the proposed multiobjective evolutionary algorithm helps to diversify the outputs across the Pareto frontier.

\section{Introducing Prompt Evolution} \label{intropromptevo}
Applications using evolutionary algorithms in digital art can be found since the 1970s \cite{b19}. Recent efforts to incorporate modern evolutionary algorithms with deep learning-based approaches for image generation have shown comparable performance to gradient-based methods \cite{b19,b20}. Given the generative model's inherent capability to produce natural images, we foresee a potential synergy with the power of evolutionary algorithms in producing diversified yet high-fidelity outputs that are simultaneously faithful to multi-dimensional user preferences. To this end, we conceptualize the idea of prompt evolution that we define as follows:



\begin{definition}[Prompt\;Evolution]\label{defevo} \emph{Let the outputs of a generative AI be conditioned on prompts. Prompt evolution then serves to impart evolutionary selection pressure and variation to the generative process of the AI, with the aim of producing multiple outputs that best satisfy target concepts/preferences implied by the prompts.}\end{definition}


Note that prompt evolution is applicable to different settings including but not limited to single-objective as well as various multi-\emph{X}\cite{b30} formulations (e.g. multi-objective, multi-task, multi-constrained) where a collection of multiple target solutions are sought.

\subsection{Relationship to Background Work}
There exists alternative techniques in the literature that aims to satisfy user preferences in generative AI, as described in what follows.

\subsubsection{Fine Tuning\cite{b31}} updates pre-trained model weights at selective layers with respect to user preferences.

\subsubsection{Prompt Tuning\cite{b27}} uses learnable ``soft prompts'' to condition frozen pre-trained language model for text-to-image generation task. 

\subsubsection{Prompt Engineering\cite{b10}} is a practice where users modify the prompts and generative model hyperparameters.

\subsubsection{In-Context Learning\cite{b26}} is an emergent behavior in large language models where a frozen language model performs a new task by conditioning on a few examples of the task.



The aforementioned techniques often generate a single output that satisfies user preferences to some extent. However, such techniques involve implicit trade-offs between producing outputs of high-fidelity, remaining faithful to the prompt, and generating diverse outputs. In contrast, prompt evolution generates multiple outputs in a single run to provide a well-balanced approach that addresses all these trade-offs simultaneously (as per Definition \ref{defevo}).

\section{Prompt Evolution with Classifier-guidance}


This section presents one instantiation of the broader idea of prompt evolution (Definition \ref{defevo}). In particular, it exemplifies how \emph{multi-objective} evolutionary selection pressure is imparted to the generative process of AI, producing \emph{diverse} images that best satisfy user preferences inferred from the prompts.

\subsection{Preliminaries}

Let  \( \mathcal{X} \subset \mathbb{R}_{\geq 0}^{\left(N \times K \times C \right)} \) denote a space of images of height $N$, width $K$ and $C$ channels, $\theta$ be the user prompt expressed in free-form natural language, \( \mathcal{L} = \{ \lambda_{1}, … , \lambda_{Q} \} \) be a set of \( Q \) user preference labels, and \( \mathcal{Y} = [0,1]^{Q} \) be a probability space where \( y_i \)  is the probability that user preference \( \lambda_{i} \) is satisfied. User preferences are simple concepts, such as actions or desired emotion descriptors, that serve to augment the prompts. 

Given any image \( \textbf{x} \in \mathcal{X} \), we assume to have multi-label classifiers \( \mathcal{F}:\mathcal{X} \rightarrow \mathcal{Y} \) that can predict the probabilities corresponding to the \( Q \) user preference labels. We also assume a conditional deep generative model \( \mathcal{G} \) from which new samples can be iteratively generated guided by the prompts as:

\begin{equation}
    \textbf{x}_{t}^{[m]} \sim \left\{
    \begin{array}{ll}
        P_{\mathcal{G}}\left(\textbf{x}_t \mid \theta \right) & t=0  \\
        P_{\mathcal{G}}\left( \textbf{x}_t \mid \theta, \textbf{x}_{t-1}^{[l]} \right) & otherwise 
    \end{array}
    \right.
\end{equation}

\noindent where \(t\)  is the iteration count such that \( 0 \leq t \leq \mathcal{T}_{max} \). Notice the salient feature that the probability distribution of the $m$-th image sample in iteration $t$ is conditioned on the $l$-th image sample at iteration $t-1$. This feature shall be uniquely leveraged as a novel mutation operator in the proposed multiobjective evolutionary algorithm for prompt evolution.

\subsection{Problem Formulation}

The $Q$-objective optimization problem statement for prompt evolution is then:

\begin{equation}
    \begin{split}
        \max_{\textbf{x}} \mathcal{F}(\textbf{x}) = \left[y_1(\textbf{x}), y_2(\textbf{x}), \cdots, y_Q(\textbf{x}) \right]    
        \\ \text{s.t. } d\left( \textbf{x}, \theta \right) \leq b.
    \end{split}
\end{equation}

\noindent Here, $b$ is a predefined upper bound constraining the deviation \( d\left( \textbf{x}, \theta \right) \) between generated outputs \( \textbf{x} \) and the user prompt \( \theta \). The deviation is computed using:

\begin{equation}
   d\left( \textbf{x}, \theta \right) = \tau \left[ 1 - \cos \left( \textbf{x}_{CLIP}, \theta_{CLIP} \right) \right]
\end{equation}

\noindent where \( \tau > 0 \) is the temperature hyperparameter and \( \textbf{x}_{CLIP} \)  and \( \theta_{CLIP} \) are the normalized CLIP \cite{b21} embeddings.

\begin{figure}[htbp]
\centerline{\includegraphics[scale=0.43]{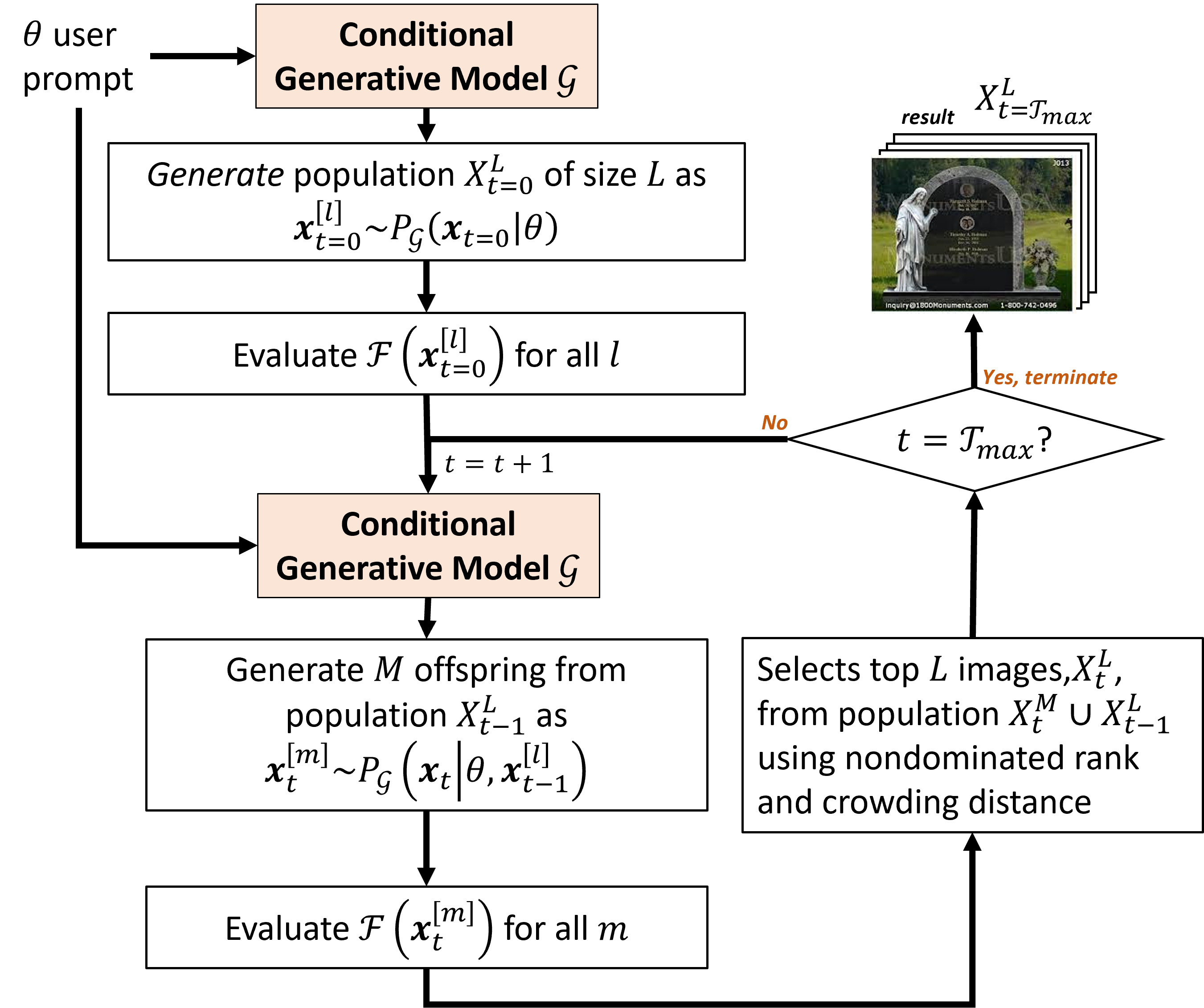}}
\caption{Workflow of prompt evolution with multi-label classifier-guidance}
\label{fig:architecture}
\end{figure}

Outputs from the multi-label classifiers serve as the fitness function of the evolving images, which imparts a selection pressure on images to better fit user preferences at every generation. The user preferences augment the prompts to reduce the disconnect between the generated images and the desired target concepts. Fig. \ref{fig:architecture} illustrates the workflow of the multiobjective evolutionary algorithm.

It is worth highlighting that standard evolutionary algorithms typically use simple probabilistic models (e.g., Gaussian distributions) as mutation operators. In contrast, the proposed algorithm in Fig. \ref{fig:architecture} makes use of the conditional generative model \( \mathcal{G} \) to implicitly serve as a kind of mutation. To our knowledge, this novelty hasn't been explored in the literature.

\section{Experimental Overview}
We explored the performance of prompt evolution in generating diverse images from prompts derived from ``inspirational proverbs''. The chosen user preferences are entities, scenes, actions, or desired emotional responses that the images are intended to evoke. These preferences are implicitly embedded in the prompt. Furthermore, we selected the NSGA-II algorithm\cite{b29} and pre-trained Stable Diffusion model\cite{b17} for our experiments. Lastly, we used pre-trained multi-label classifiers to predict the expected entities, concepts, and emotional responses \cite{b23,b24,b25}.

\section{Experimental Results}
For each set of text prompts describing a proverb, we identified 2 to 3 user preferences as the multiple objectives to optimize. We ran multiple runs using the same set of text prompts, with each run completing 20 generations of evolution. The pre-trained generative model parameters are frozen throughout all our experiments.

In our results, before prompt evolution, we observed that the pre-trained generative model often failed to depict the multiple target concepts and their dependencies. Hence, the model failed to generate images aligned with user preferences (see Table \ref{table:1}). Through evolution, the method was able to fit these preferences increasingly well, producing better variations and spreading across the Pareto frontier with every new generation (see Figure \ref{fig:insight}). As a result, the evolved images could faithfully capture expected user preferences.

\subsection{Comparison Study}
\begin{figure}[htbp]
\centerline{\includegraphics[scale=0.24]{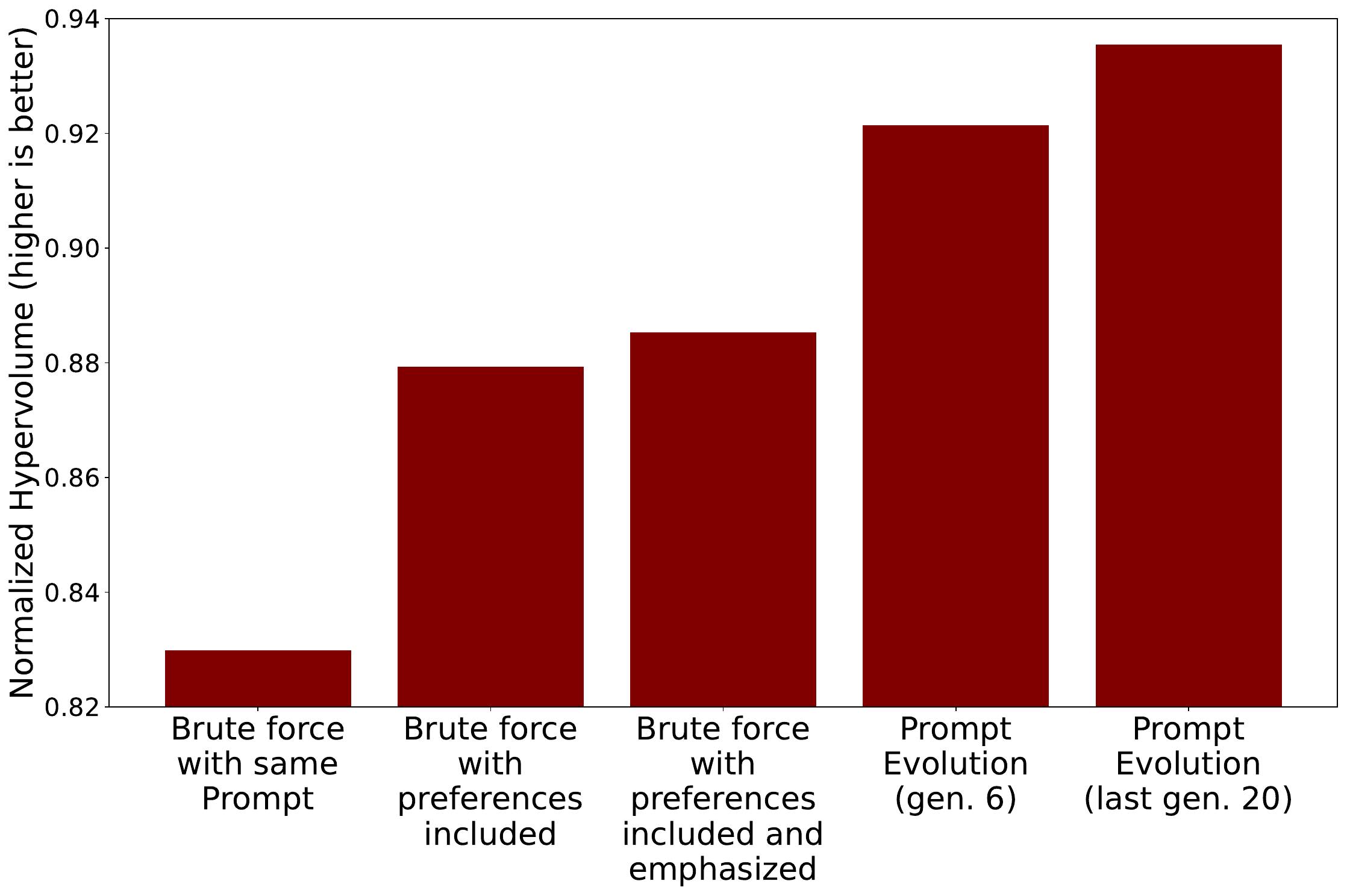}}
\caption{Comparison of brute force approach versus Prompt Evolution with multi-label classifier-guidance}
\label{fig:comparison}
\end{figure}

\begin{figure*}
  \centering
  \includegraphics[scale=0.42]{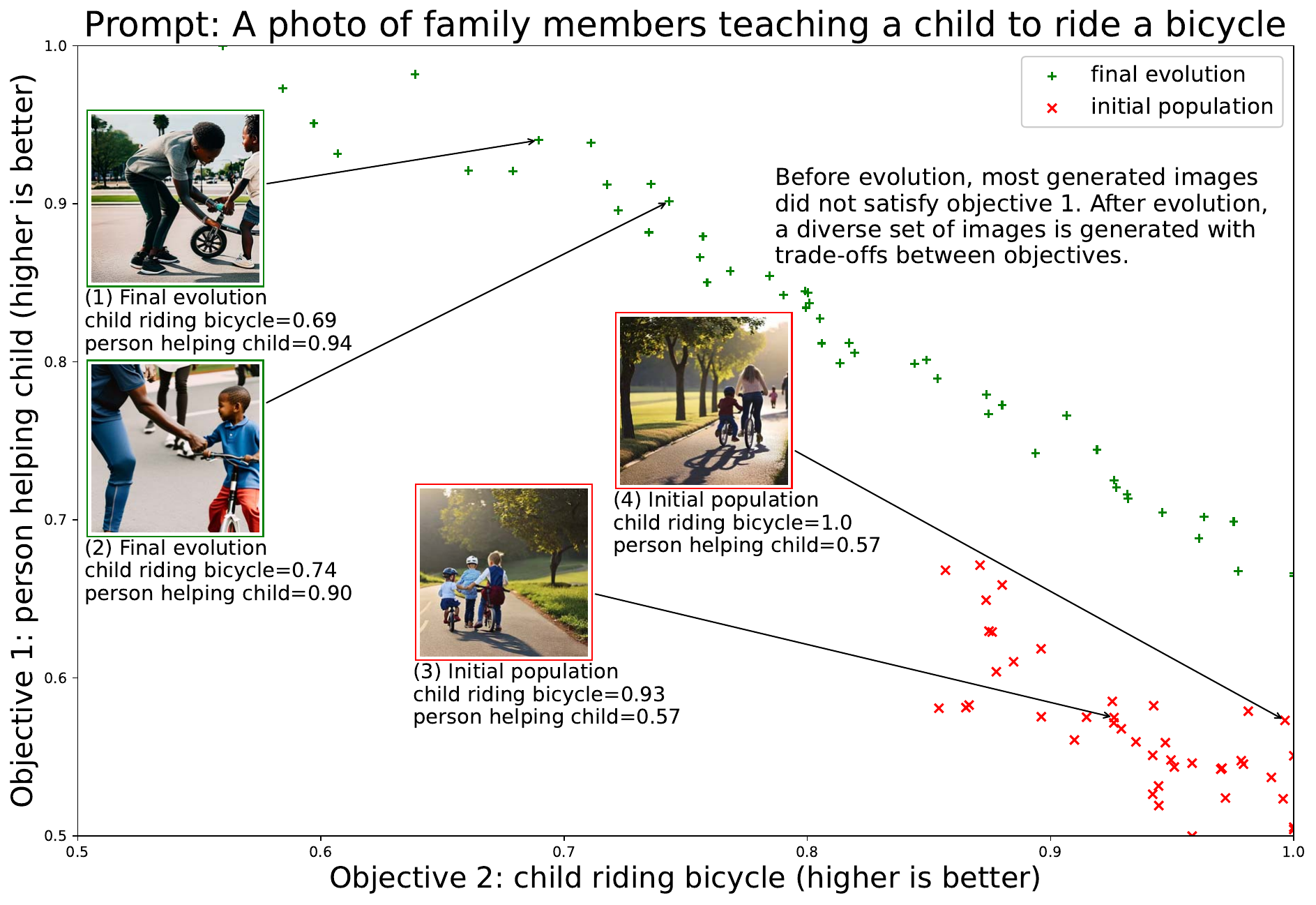}
  \caption{Prompt derived from the proverb: ``The family that plays together stays together". The nondominated population of generated images at the initial and final stages of evolution is shown. Before evolution, most generated images did not satisfy objective 1. After evolution, a diverse set of images is generated with trade-offs between objectives.}
  \label{fig:insight}
\end{figure*}

We extend the experiment presented in Figure \ref{fig:insight} to compare the diversity of generated images between prompt evolution and a brute-force approach, with and without prompt engineering, using hypervolume\cite{b28} as the performance metric. The results are presented in Figure \ref{fig:comparison}. Note that the brute force approach generates the same number of images as prompt evolution has sampled from the same pre-trained generative model. 

As shown by Figure \ref{fig:comparison}, prompt evolution outperforms the brute-force approach as early as generation 6. This demonstrates that prompt evolution is indeed efficient in finding diversified and near Pareto optimal images. Moreover, we observed that while prompt engineering in the brute force approach generates some images more faithful to the prompt, improvements in generating diversified images were marginal - even for a considerable sample size of 1550 images.

\begin{table}[h!]
\centering
\begin{tabular}{||c c c ||} 
\hline
 \multicolumn{3}{|l|}{Proverb: Practice makes perfect} \\
 \multicolumn{3}{|l|}{Prompt: A sketch of a person diligently studying a book under a lamp} \\
 \multicolumn{3}{|l|}{Objectives: ``person reading book", ``lamp above person", ``awe"}\\ [0.5ex] 
 \hline
 Before evolution & \multicolumn{2}{|c|}{After prompt evolution} \\ [1ex]  
 \includegraphics[scale=0.08]{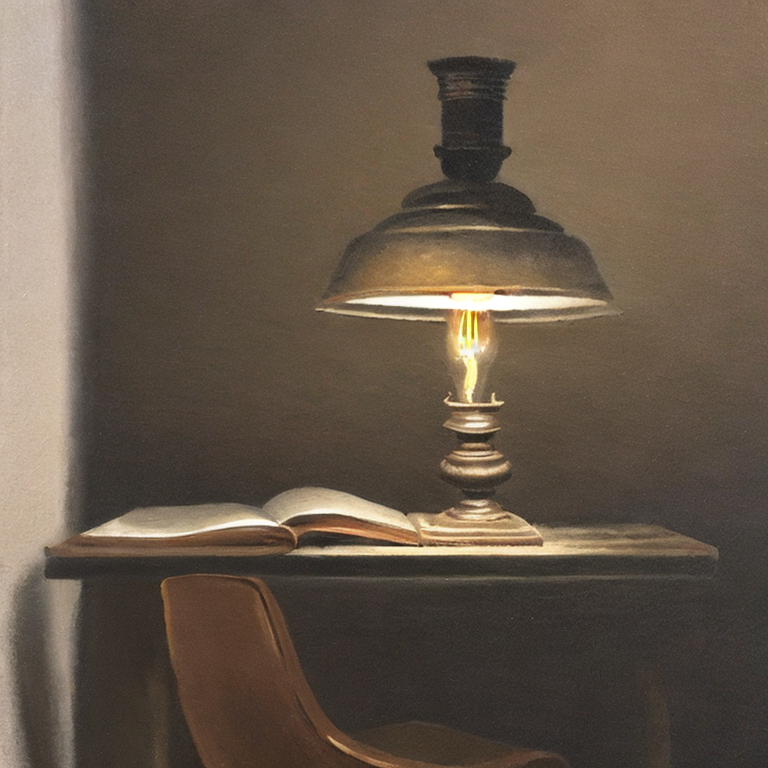} &
 \multicolumn{2}{|l|}{ 
 \includegraphics[scale=0.08]{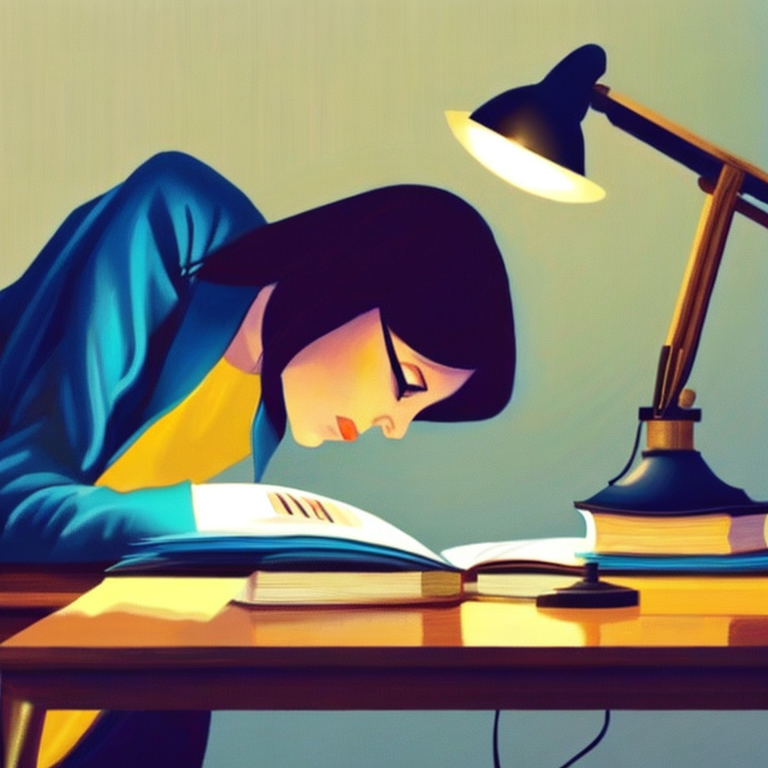}
 \includegraphics[scale=0.08]{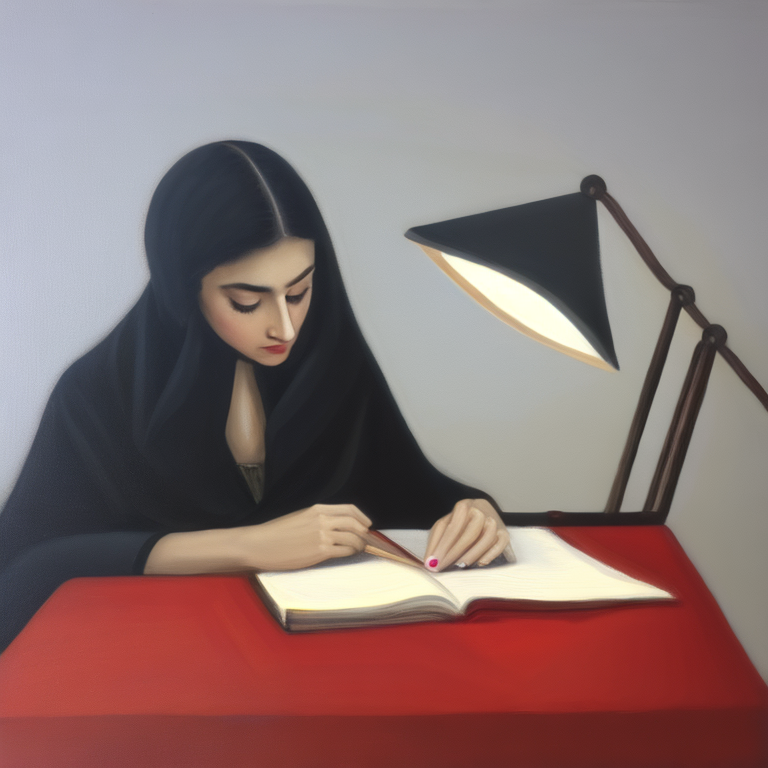}
 \includegraphics[scale=0.08]{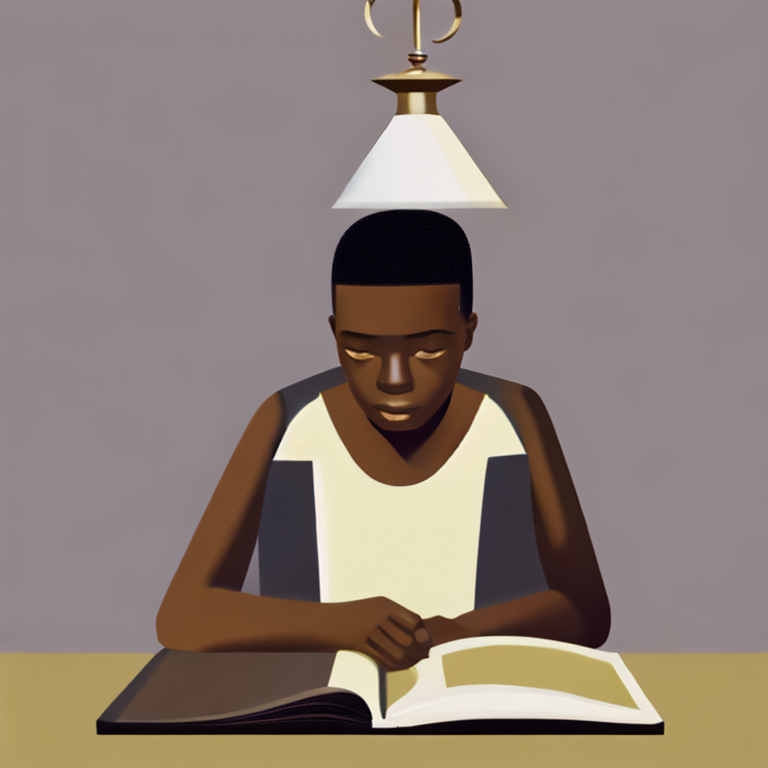}
 } \\ [1ex] 
 \hline
 \multicolumn{3}{|c|}{(a)} \\ [1ex]
 \hline\hline
 \multicolumn{3}{|l|}{Proverb: A journey of thousand miles begins with a single step} \\
 \multicolumn{3}{|l|}{Prompt: A person standing at the edge of a cliff with a backpack and} \\
 \multicolumn{3}{|l|}{\hspace{1cm}looking into the distance} \\
 \multicolumn{3}{|l|}{Objectives: ``person standing on cliff", ``person carrying backpack", ``excitement"}\\ [0.5ex] 
 \hline
 Before evolution & \multicolumn{2}{|c|}{After prompt evolution} \\ [1ex]  
 \includegraphics[scale=0.08]{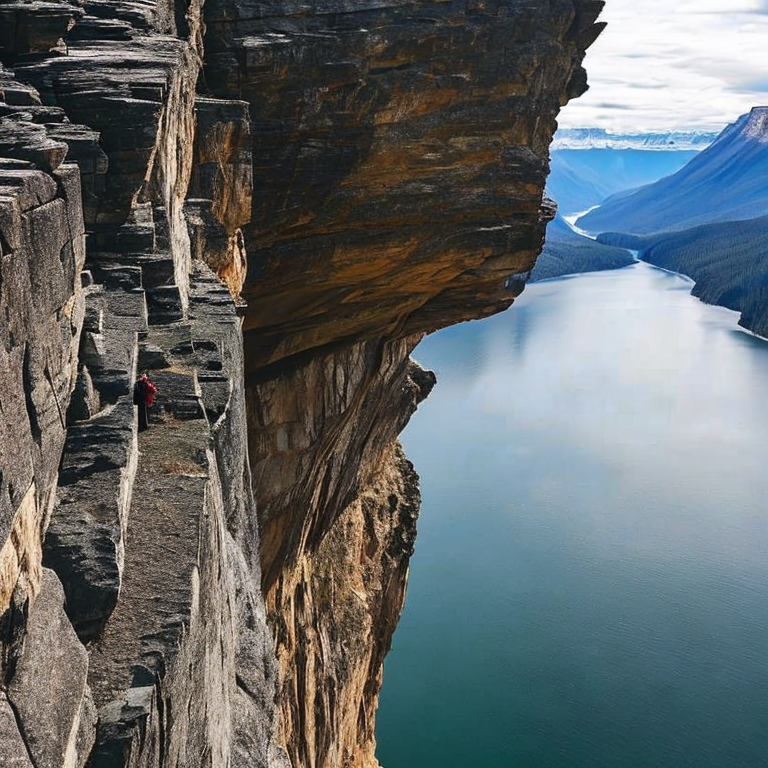} &
 \multicolumn{2}{|l|}{ 
 \includegraphics[scale=0.08]{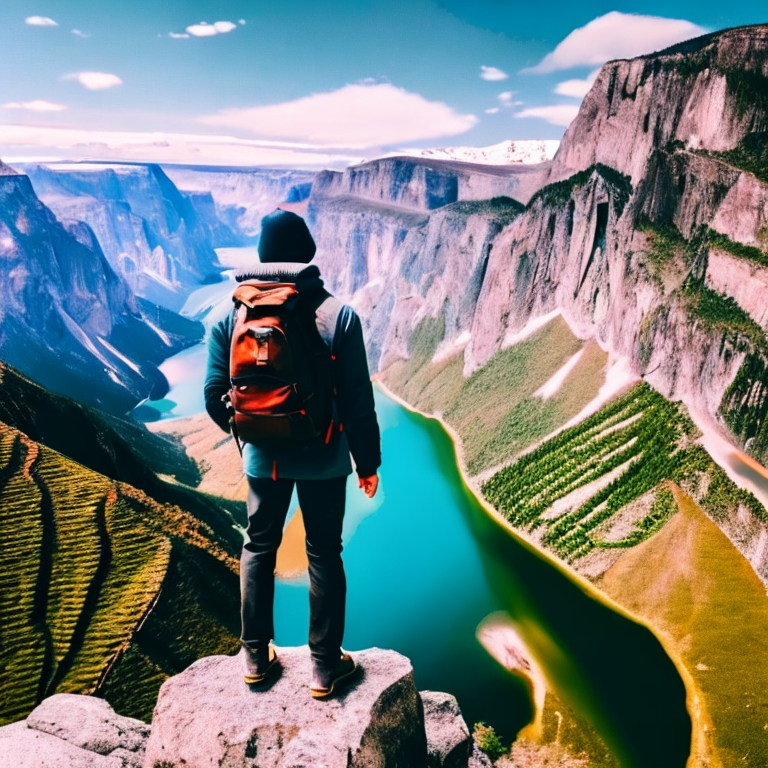}
 \includegraphics[scale=0.08]{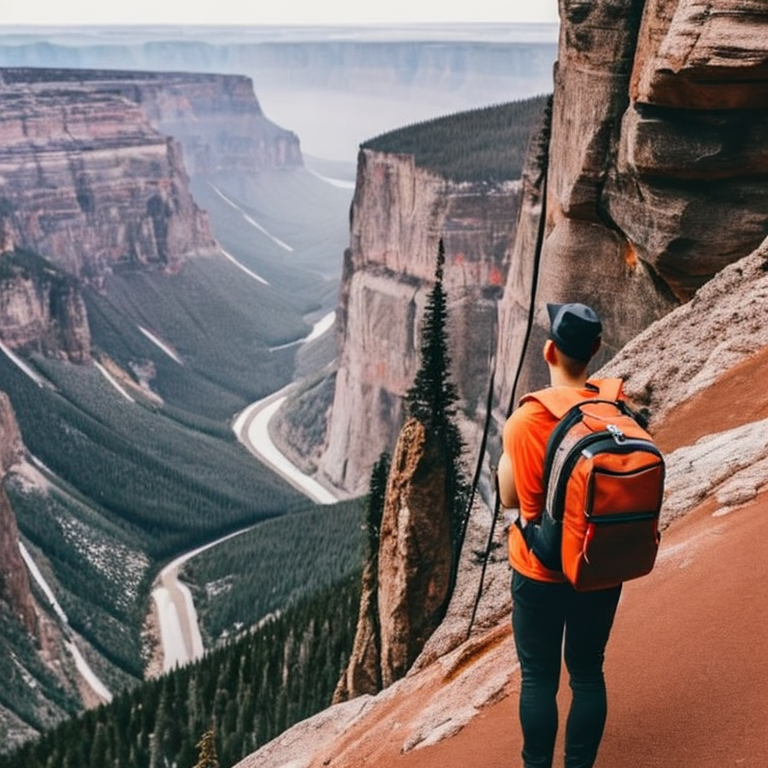}
 \includegraphics[scale=0.08]{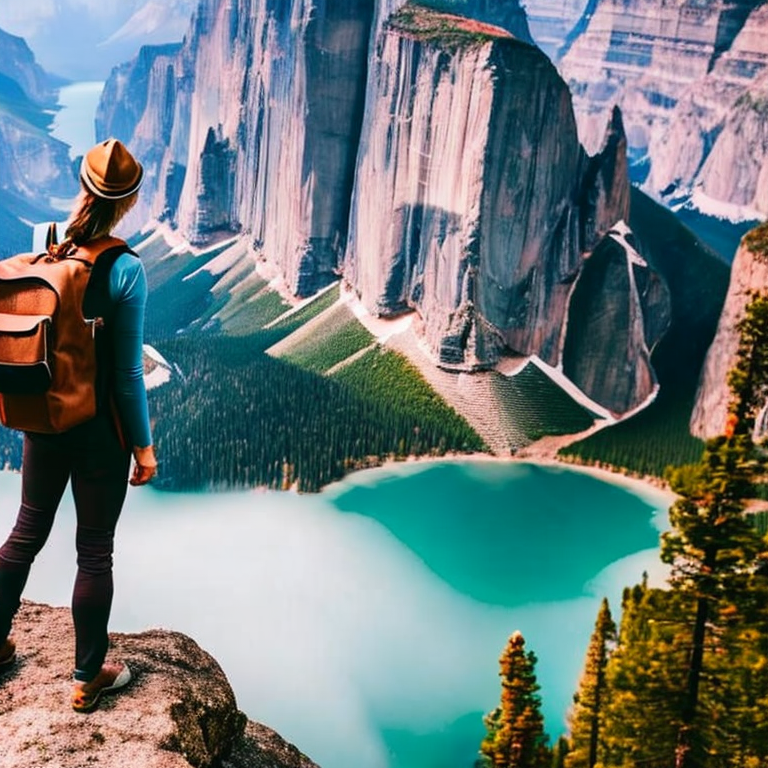}
 } \\ [1ex] 
 \hline
 \multicolumn{3}{|c|}{(b)} \\ [1ex]
 \hline\hline
\end{tabular}
\caption{Examples of images generated before and after prompt evolution with multi-label classifier-guidance}
\label{table:1}
\end{table}


\section{Conclusions and future research}
This paper presented prompt evolution as a novel, automated approach to creating images faithful to user preferences. The method synergizes the inherent ability of generative models to produce natural images with the power of diversification of evolutionary algorithms. Considerable improvements in the quality of images are achieved, alleviating shortcomings of today's deep models. As highlighted in Section \ref{intropromptevo}, prompt evolution need not be limited to multi-objective settings and can also apply to other settings such as single-objective or various multi-\emph{X} formulations. Additionally, prompt evolution exerts evolutionary selection pressure and variation not only to the generated outputs but also to other forms of guidance and hidden states such as prompts, audio, and latent embeddings, to name a few. Such flexibility within the paradigm indeed presents abundant research opportunities in this emerging field of prompt evolution.




\end{document}